\documentclass[10pt,twocolumn,letterpaper]{article}

\usepackage{iccv}
\usepackage{times}
\usepackage{epsfig}
\usepackage{graphicx}
\usepackage{amsmath}
\usepackage{amssymb}
\usepackage{booktabs}
\usepackage{makecell}

\usepackage[breaklinks=true,bookmarks=false]{hyperref}
\usepackage[capitalize]{cleveref}
\iccvfinalcopy 

\def\iccvPaperID{****} 

\ificcvfinal\fi

\begin{document}

\title{APPT: Asymmetric Parallel Point Transformer for 3D Point Cloud Understanding}

\author{Hengjia Li\\
Zhejiang University\\
\and
Tu Zheng\\
Fabu\\
\and
Zhihao Chi\\
Zhejiang University\\
\and
Zheng Yang\\
Fabu\\
\and
Wenxiao Wang\\
Zhejiang University\\
\and
Boxi Wu\\
Zhejiang University\\
\and
Binbin Lin\\
Zhejiang University\\
\and
Deng Cai \\
Zhejiang University\\
}

\maketitle

\begin{abstract}
Transformer-based networks have achieved impressive performance in 3D point cloud understanding. 
However, most of them concentrate on aggregating local features, but neglect to directly model global dependencies, which results in a limited effective receptive field. 
Besides, how to effectively incorporate local and global components also remains challenging.
To tackle these problems, we propose \textbf{Asymmetric Parallel Point Transformer} (APPT). 
Specifically, we introduce \textbf{Global Pivot Attention} to extract global features and enlarge the effective receptive field. 
Moreover, we design the \textbf{Asymmetric Parallel} structure to effectively integrate local and global information. 
Combined with these designs, APPT is able to capture features globally throughout the entire network while focusing on local-detailed features.
Extensive experiments show that our method outperforms the priors and achieves state-of-the-art on several benchmarks for 3D point cloud understanding, such as 3D semantic segmentation on S3DIS, 3D shape classification on ModelNet40, and 3D part segmentation on ShapeNet. 
\end{abstract}

\section{Introduction}
\label{sec:intro}
3D point cloud understanding has been a fundamental task in various application areas, \eg, autonomous driving, robotics, and AR/VR. 
Structurally different from images, point clouds are sets in the continuous 3D space inherently, which are sparse, irregular, and unordered. 
These properties make it difficult to apply the mature deep network in 2D image understanding to 3D point cloud understanding directly.
\begin{figure}
\begin{center}

	\centering
    \begin{minipage}  {0.32\linewidth}
        \centering
        \includegraphics [width=0.9\linewidth,height=0.8\linewidth]
        {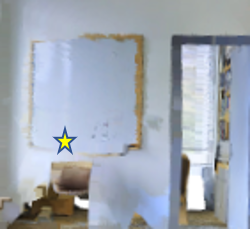}
    \end{minipage}      
    \begin{minipage}  {0.32\linewidth}
        \centering
        \includegraphics [width=0.9\linewidth,height=0.8\linewidth]
        {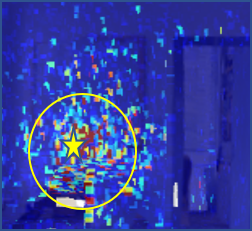}
    \end{minipage}      
     \begin{minipage}  {0.32\linewidth}
        \centering
        \includegraphics [width=0.9\linewidth,height=0.8\linewidth]
        {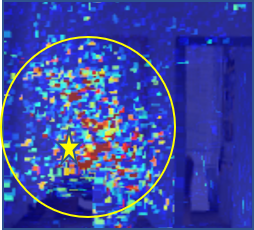}
    \end{minipage} 
	 
	 
    \begin{minipage}  {0.32\linewidth}
        \centering
        \includegraphics [width=0.9\linewidth,height=0.8\linewidth]
        {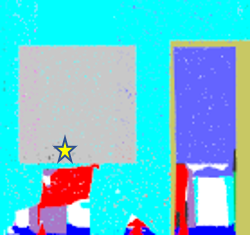}\\\footnotesize Input / Ground Truth
    \end{minipage}      
    \begin{minipage}  {0.32\linewidth}
        \centering
        \includegraphics [width=0.9\linewidth,height=0.8\linewidth]
        {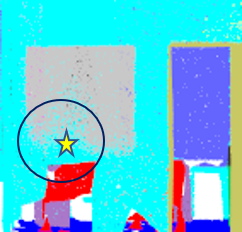}\\\footnotesize w/o global branch (PT)
    \end{minipage}      
     \begin{minipage}  {0.32\linewidth}
        \centering
        \includegraphics [width=0.9\linewidth,height=0.8\linewidth]
        {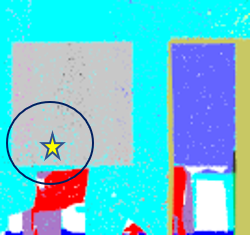}\\\footnotesize w/ global branch (ours)
    \end{minipage} 
    
    \vspace{0.1cm}

    \begin{minipage}  {0.06\linewidth}
        \centering
        \includegraphics [width=0.5\linewidth,height=0.5\linewidth]
        {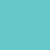}
    \end{minipage}\footnotesize board
    \begin{minipage}  {0.06\linewidth}
        \centering
        \includegraphics [width=0.5\linewidth,height=0.5\linewidth]
        {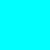}
    \end{minipage}\footnotesize wall
    \begin{minipage}  {0.06\linewidth}
        \centering
        \includegraphics [width=0.5\linewidth,height=0.5\linewidth]
        {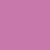}
    \end{minipage}\footnotesize table
    \begin{minipage}  {0.06\linewidth}
        \centering
        \includegraphics [width=0.5\linewidth,height=0.5\linewidth]
        {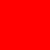}
    \end{minipage}\footnotesize chair
        \begin{minipage}  {0.06\linewidth}
        \centering
        \includegraphics [width=0.5\linewidth,height=0.5\linewidth]
        {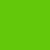}
    \end{minipage}\footnotesize door
        \begin{minipage}  {0.06\linewidth}
        \centering
        \includegraphics [width=0.5\linewidth,height=0.5\linewidth]
        {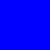}
    \end{minipage}\footnotesize floor
        \begin{minipage}  {0.06\linewidth}
        \centering
        \includegraphics [width=0.5\linewidth,height=0.5\linewidth]
        {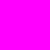}
    \end{minipage}\footnotesize window
\end{center}
\vspace{-0.2cm}
\caption{Visualization of Effective Receptive Field (ERF)~\cite{luo2016understanding}. The star is the point of interest. The yellow circle in the top row is the size of ERF and the black circle in the bottom row corresponds to the comparison of predictions.
\textbf{Left}: Input point cloud and the ground truth. \textbf{Middle}: ERF and prediction of the model without global branch, similarly to Point Transformer (PT)\cite{zhao2020point}. \emph{The ERF is limited to the neighbors of the star which makes it mistakenly identify the board as the wall}. \textbf{Right}: ERF and prediction of the ours with global branch. 
\emph{Our ERF is larger which helps to correct the prediction.}}
\label{fig:erf}
\vspace{-0.4cm}
\end{figure}

Transformer, which becomes popular in both natural language processing and 2D image understanding, is particularly suitable for 3D point cloud understanding. It is invariant to the permutation and cardinality of the input elements\cite{zhao2020point}, which makes it a set operator inherently.
Motivated by the advantages, several works\cite{zhao2020point,XinLai2022StratifiedTF,zhang2022patchformer,ChunghyunPark2022FastPT} attempt to develop the Transformer-based modules.
However, due to the quadratic complexity of the attention operator, previous works still struggle to build the global context for 3D point cloud understanding.

As a pioneering work, Point Transformer\cite{zhao2020point} adopts Local Group Attention (LGA) to aggregate features from grouped neighbors but neglects to explicitly capture long-range contexts, which results in the limited effective receptive field.
As a result, Point Transformer fails to predict the \textit{board}
accurately as shown in the middle of \cref{fig:erf}.
Recently, Stratified Transformer\cite{XinLai2022StratifiedTF} adopts window-based self-attention and proposes a key sampling strategy to partially tackle this problem. 
However, its computation cost increases significantly compared with Point Transformer and \emph{the effective receptive field is still limited} due to the window-based attention which loses the non-locality \cite{ZhengzhongTu2022MaxViTMV}.

Another issue is how to effectively incorporate the local and
global components. 
Local features convey fine geometric information\cite{qi2017pointnet++}, \eg, the texture and margin while global features are with more semantic meanings.
Thus previous works\cite{nie2022pyramid,sun2019high,tan2020efficientdet} propose feature integration for boosting performance. 
These methods motivate us that both local and global information is complementary for 3D point cloud understanding, which leaves a problem that \emph{how to effectively integrate local and global features across the network.}

To enlarge the effective receptive field, we propose novel 
\textbf{Global Pivot Attention} (GPA) which directly models the long-range dependencies.
Considering that adopting the vanilla self-attention across all points will bring nonnegligible computation costs due to millions of points, we utilize the Farthest
Point Sampling algorithm\cite{qi2017pointnet++} to sample the input points and get some evenly distributed points, termed as the \textbf{pivot points}.
Through the pivot points, GPA is able to promote the interaction between all points and capture long-range dependencies effectively.
On the other hand, instead of vector attention utilized in LGA, GPA adopts scalar attention which also reduces the computation cost significantly.  

Moreover, we introduce \textbf{parallel} branches to extract local and global features respectively.
Specifically, we utilize GPA as the global branch and LGA within grouped neighbors as the local branch, which also introduces no additional overhead compared with Point Transformer.
Besides, we design the \textbf{asymmetric} structure to effectively incorporate local and global information.
As is observed in \cite{MaithraRaghu2021DoVT,si2022inception}, the bottom layers usually tend to capture the local information such as textures while the top layers pay more attention to the global information such as the structures of the whole scene. 
Hence, we adopt the ramp channel-ratio structure that gradually increases the channel-ratio of the input feature delivered into the global branch from the bottom to the top.
In the meanwhile, 
the ratio of pivot points in the global branch is gradually increased in a similar way.

To verify the effectiveness of these designs, we further build a framework for 3D point cloud understanding, named \textbf{Asymmetric Parallel Point Transformer} (APPT).
As shown in the right of \cref{fig:erf}, the point can interact with further points of the \textit{board}, which helps to aggregate global contexts and correct the prediction, indicating that our global branch has the powerful capability to model the long-range dependencies and enlarge the effective receptive field.
Furthermore, experimental results demonstrate that our model outperforms prior works and sets the new state-of-the-art on various benchmarks for 3D point cloud understanding, \eg, 3D semantic segmentation on S3DIS, 3D
shape classification on ModelNet40 and 3D part segmentation on ShapeNet.

In summary, our contributions are as follows:
\begin{itemize}
    \item We propose a novel transformer block for point cloud understanding, which effectively models short-range and long-range dependencies via parallel branches. Moreover, we introduce Global Pivot Attention (GPA) as the global branch that enlarges the effective receptive field. \emph{To our knowledge, this is the first work to attempt full attention for point cloud understanding.}  
    \item To effectively integrate local and global information, we design the Asymmetric Parallel structure that increases the channel-ratio and sampling-ratio of the global branch from the bottom layers to the top layers.
    \item Based on these designs, we build the network for classification and segmentation on 3D point clouds, termed as Asymmetric Parallel Point Transformer (APPT). Experimental results demonstrate that our method outperforms previous state-of-the-art methods and introduces no additional computation costs.
\end{itemize}

\section{Related Work}
\label{sec:related}
In this section, we first revisit vision transformers in 2D image understanding. Then we review previous methods for 3D point cloud understanding and transformer-based methods specifically.

\noindent
{\bf Vision Transformer.}
Motivated by the success of transformers in natural language processing\cite{vaswani2017attention}, ViT\cite{dosovitskiy2020vit} adopts the pure-transformer architecture in 2D image understanding and the competitive performance shows its outstanding ability to model the long-range dependencies. 
However, the pure-transformer architecture is inefficient to encode local features\cite{NamukPark2022HowDV} and its computational complexity is quadratic to the token numbers. 
Recently, many variants are designed to improve the locality\cite{liu2021swin,si2022inception,YunhaoWang2022MAFormerAT,ZhengzhongTu2022MaxViTMV,WenxiaoWang2021CrossFormerAV,RuiYang2022ScalableViTRT,SuchengRen2021ShuntedSV} and computational complexity\cite{YouweiLiang2022NotAP,MichaelSRyoo2021TokenLearnerWC, fan2021multiscale}. 
To account for the locality, 
Inception Transformer\cite{si2022inception} splits the channels and adopts parallel branches to combine self-attention and convolution.
On the other hand, MViT\cite{fan2021multiscale} scales the spatial channel and creates a multiscale feature pyramid to reduce computation cost. 

\noindent
{\bf Point Cloud Understanding.}
Generally, methods for 3D point cloud understanding are classified into two categories, \ie voxel-based and point-based methods.

The voxel-based methods\cite{choy20194d,3DSemanticSegmentationWithSubmanifoldSparseConvNet,JiagengMao2021VoxelTF} first voxelize the 3D space, 
and then adopts 3D convolution on the voxels. Recently, Mao \etal \cite{JiagengMao2021VoxelTF} proposes a voxel-based transformer, which adopts both local and dilated attention to capture multi-scale features. Despite having decent performance, it inevitably loses the geometric details of the original point cloud due to voxelization.
Besides,  it has ‘gridding effect` due to the dilated architecture and the voxel structure amplifies this effect when point cloud is
dense and the voxel has plenty of points which often occurs
in the indoor scenarios.

Instead of voxelization, the point-based methods\cite{qi2017pointnet,qi2017pointnet++,thomas2019kpconv,wu2019pointconv,guo2021pct,zhao2020point,XinLai2022StratifiedTF,choe2022pointmixer, qianpointnext, kaul2022convolutional} directly adopt the original points feature as inputs.
As the pioneering work, PointNet\cite{qi2017pointnet} utilizes MLP and pooling to process features. 
Based on PointNet, PointNet++\cite{qi2017pointnet++} adopts hierarchical sampling strategies to extract local features.
Specifically, it adopts the Farthest Point Sampling algorithm to down-sample the points and the K-Nearest Neighbor algorithm to search for local neighborhood points. 
Subsequent to PointNet and PointNet++, many point-based methods are developed that mainly focus on proposing novel modules to capture local features.
For example, KPConv\cite{thomas2019kpconv} proposes the pseudo-grid convolution using kernel points which mimics the convolution in 2D image understanding. 
However, most of them neglect the modeling of long-range dependencies, leading to a limited effective receptive field which hinders their potential for point cloud understanding.

\noindent
{\bf Transformers in Point Cloud Understanding.}
Equipped with self-attention, Transformer has great power to capture long-range dependencies. 
Some attempts\cite{zhao2020point,XinLai2022StratifiedTF,ChunghyunPark2022FastPT,zhang2022patchformer, wupoint} have been made to develop the Transformer-based module. 
As a pioneering work, Point Transformer\cite{zhao2020point} adopts local group attention to aggregate features and subtraction relation to generate the attention weights, but it still lacks the module that directly captures long-range dependencies.

Recently, Stratified Transformer\cite{XinLai2022StratifiedTF} proposes a transformer block based on Swin\cite{liu2021swin} and a novel key sampling strategy to enlarge the effective receptive field.
Despite achieving high performance in 3D point cloud segmentation, it still has a limited effective receptive field due to the window-based attention which lacks the non-locality. Besides, it significantly increases the computation costs compared with Point Transformer\cite{zhao2020point}.

Based on Point Transformer, we propose the novel Asymmetric Parallel Point Transformer without additional computation costs, which adopts Global Pivot Attention as the global branch to capture long-range dependencies and further enlarge the effective receptive field.

\begin{figure*}[h]

	\centering
	\includegraphics[width=0.8\linewidth]{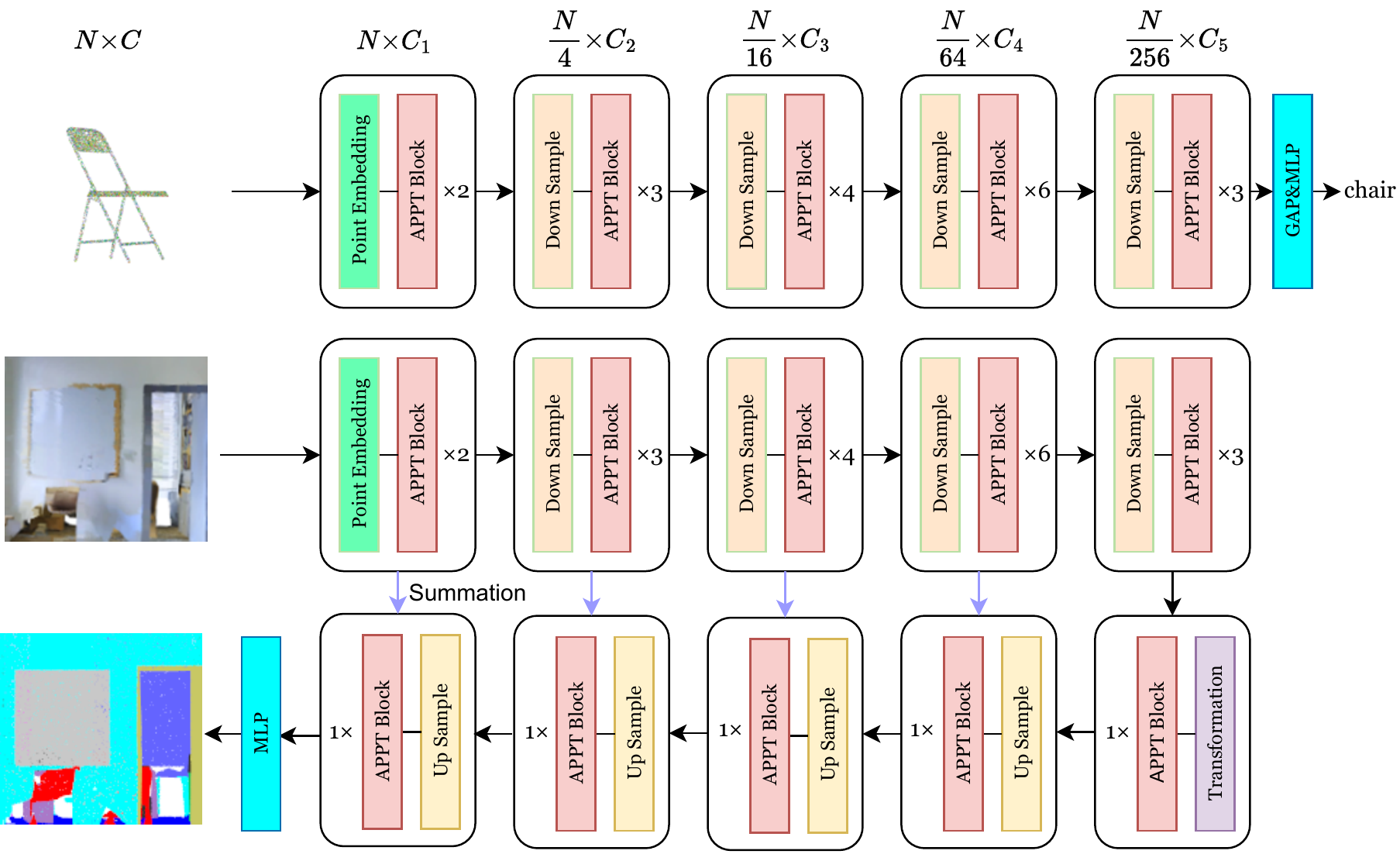}

	\caption{Structure of Asymmetric Parallel Point Transformer. For 3D shape classification on the top row, the input features are firstly embedded via MLPs and go through APPT blocks. After several stages composed of the down-sample layer and APPT blocks, the features are delivered into the classification head which consists of Global Average Pooling and MLPs.
    For 3D semantic segmentation on the bottom rows, the Encoder is the same as the 3D shape classification network. After that, the features are fed into the stages composed of the up-sample layer and one APPT block. Then the features go through the classification head composed of MLPs. The black arrow is the flow of features and the purple arrow is the summation in the up-sample layer.}
	\label{fig:network}
	
\end{figure*}

\section{Method}
\label{sec:method}
In this section, we first revisit local group attention in Point Transformer\cite{zhao2020point} with a brief review of standard self-attention in pure transformers. Then we introduce our proposed Asymmetric Parallel Point Transformer in detail.
\subsection{Revisit Point Transformer}
\noindent
{\bf Standard Self-Attention.}
Transformers with standard self-attention\cite{vaswani2017attention,dosovitskiy2020vit} have achieved impressive performance in both natural language processing and 2D image understanding, benefiting from their significant ability to capture long-range dependencies. Let $\mathcal{X}=\{\mathbf{x}_i\}_{i=1}^{N}$ be a set of feature. The standard scalar self-attention in the pure transformer can be represented as follows:
\begin{equation}
\label{standard}
\begin{aligned}
    \mathbf{Q}_i = \varphi(\mathbf{x}_i),\mathbf{K}_j &= \psi(\mathbf{x}_j),\mathbf{V}_j = \alpha(\mathbf{x}_j),\\
    \mathbf{A}_{i, j} &= \gamma(\mathbf{Q}_i \mathbf{K}_j^\top),\\
    \mathbf{y}_i = &\sum_{\mathbf{x}_j \in \mathcal{X}}\rho(\mathbf{A}_{i, j}) \mathbf{V}_j,
\end{aligned}
\end{equation}
where $\mathbf{y}_i$ is the output feature, $\varphi$, $\psi$, $\alpha$ and $\gamma$ are MLPs. $\rho$ is the softmax function. 

\noindent
{\bf Local Group Attention.}
In point cloud understanding, assume a set of point cloud $\mathcal{P}=\{\mathbf{p}_i\}_{i=1}^{N}$ and the corresponding feature $\mathcal{X}=\{\mathbf{x}_i\}_{i=1}^{N}$, where $N$ is the number of points. $\mathbf{p}_i$ is the position of the $i$-th point and $\mathbf{x}_i$ is the corresponding feature.

Point Transformer \cite{zhao2020point} adopts local group attention to process $\mathcal{X}$. Specifically, 
\begin{equation}
\label{ptr}
\begin{aligned}
    \delta_{i,j} &= \theta(\mathbf{p}_i - \mathbf{p}_j),\\
    \mathbf{A}_{i, j} &= \gamma(\mathbf{Q}_i - \mathbf{K}_j + \delta_{i,j}),\\
    \mathbf{y}_i = &\sum_{\mathbf{x}_j \in \mathcal{X}(i)}\rho(\mathbf{A}_{i, j}) \odot (\mathbf{V}_j + \delta_{i,j}),
\end{aligned}
\end{equation}
where $\mathcal{X}(i) \in \mathcal{X}$ is the neighbor point set of the $i$-th point grouped by the K-Nearest Neighbor algorithm.
$\delta_{i,j}$ is the relative position encoding between the $i$-th and the $j$-th point.
$\theta$ and $\gamma$ are MLPs. 
$\rho$ is the Softmax and $\odot$ is the Hadamard product. 
$\mathbf{Q}_i$, $\mathbf{K}_j$, and $\mathbf{V}_j$ is the same as \cref{standard}.

While Point Transformer with local group attention shows its strong capability to extract local features, the effective receptive field is limited to grouped neighbors.
For example, when $k=16$, Point Transformer extracts the features from only 16 neighbor points, which makes it inefficient to model the long-range dependencies.
On the other hand, as shown in \cref{ptr}, Point Transformer adopts vector attention to modulate channels, while with more computation costs than scalar attention.
\subsection{Asymmetric Parallel Point Transformer}
\label{APPT}
To further enlarge the effective receptive field, we propose the Asymmetric Parallel Point Transformer, termed as APPT.
As is depicted in \cref{fig:ppt}, the token mixer of APPT blocks is composed of two parallel branches which respectively extract local and global features to model short-range and long-range dependencies.

Specifically, APPT first divides the input feature into two tokens along the channel dimension and then delivers them to each branch respectively. 
In detail, the input feature of points $\mathbf{X}  \in \mathbb{R}^{N \times C} $ is split into $\mathbf{X}_l \in \mathbb{R}^{N \times C_l}$ and  $\mathbf{X}_g \in \mathbb{R}^{N \times C_g}$, where $C_l + C_g = C$. Then $\mathbf{X}_l$ and $\mathbf{X}_g$ are respectively fed into each branch. 
Specifically, the local branch adopts local group attention (LGA) utilized in Point Transformer, as is represented in \cref{ptr}, to aggregate short-range information within grouped neighbors. 
On the other hand, the long-range dependencies are modeled by the global branch,
which utilizes the novel Global Pivot Attention, depicted in \cref{GPA}. 

The output $\mathbf{Y}_l$ and $\mathbf{Y}_g$ are fused by the fusion module. Here, we adopt simple but effective concatenation and the ablation study in \cref{table:fusion} shows its effectiveness. 
Then the final output $\mathbf{Y}$ is projected via Feed Forward Network (FFN):
\begin{equation}
\label{output}
\begin{aligned}
    \mathbf{\hat{Y}} &= \mathbf{X} + \mathrm{Concat}(\mathbf{Y}_l, \mathbf{Y}_g),\\
    \mathbf{Y} &= \mathbf{\hat{Y}} + \mathrm{FFN}(\mathbf{\hat{Y}}).
\end{aligned}
\end{equation}

\subsection{Global Pivot Attention}
\label{GPA}
Considering the heavy computation cost of standard self-attention represented as \cref{standard}, due to its quadratic complexity with respect to the number of input points, we propose Global Pivot Attention (GPA) to model long-range dependencies. 
Specifically, we first utilize the Farthest
Point Sampling algorithm (FPS)\cite{qi2017pointnet++} to sample the input points and get some evenly distributed points, termed as pivot points. 
Then key ($\mathbf{K}$) and value ($\mathbf{V}$) in self-attention are mapped from pivot points, while query ($\mathbf{Q}$) is projected from all the points to maintain the resolution.
Note that we adopt scalar attention instead of vector attention to reduce computation cost and focus on the relation between points instead of channels.

Technically, given the input points $\mathcal{P}=\{\mathbf{p}_i\}_{i=1}^{N}$, pivot points $\mathcal{P}_{pivot}=\{\mathbf{p}_j\}_{j=1}^{M}$ are sampled from $\mathcal{P}$:
\begin{equation}
\label{fps}
\begin{aligned}
    SR &= M / N, \\
    \mathcal{P}_{pivot} &= \mathrm{FPS}(\mathcal{P}, SR),
\end{aligned}
\end{equation}
 where $M$ is the number of $\mathcal{P}_{pivot}$ and $SR$ is the ratio of sampling.

\begin{figure*}
        
        \centering
        \begin{minipage}[t]{.33\textwidth}
          \centering
          \includegraphics[width=.95\linewidth]{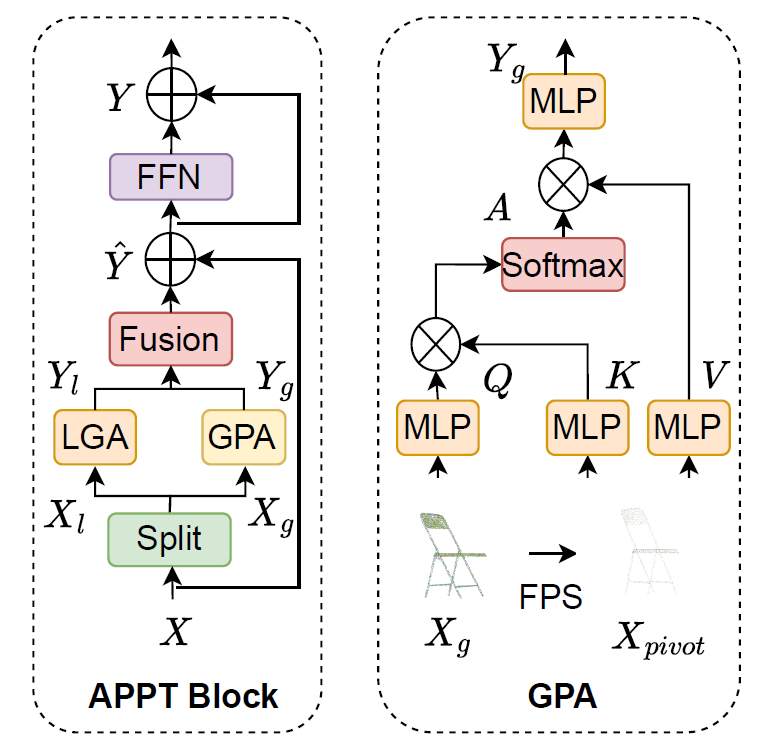}
          \caption{Details of APPT Block and GPA. The features are firstly split into two tokens and fed into Local Group Attention (LGA) and Global Pivot Attention (GPA) respectively.
          After that, the outputs are fused together and the result goes through the FFN layer, which is composed of MLPs.
          }
          \label{fig:ppt}
        \end{minipage}\hspace{0.5em}
        \begin{minipage}[t]{.63\textwidth}
          \centering
          \includegraphics[width=.9\linewidth]{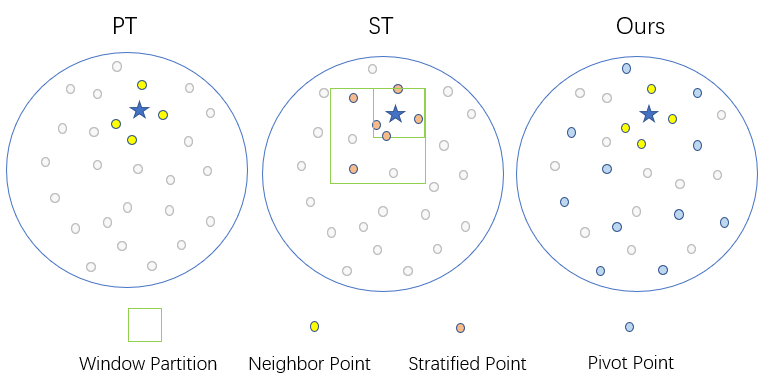}
          \caption{
            Comparison between attention in Point Transformer (PT) \cite{zhao2020point}, Stratified Transformer (ST) \cite{XinLai2022StratifiedTF} and ours.
            PT adopts local group attention within grouped neighbors to extract local features.
            ST utilizes window-based attention and a stratified key sampling strategy to partially enlarge the effective receptive field.
            Differently, we propose parallel branches to capture local and global features respectively.
            Moreover, through pivot points, the global branch directly models the long-range dependencies and further enlarges the effective receptive field.
          }
          \label{fig:GPA}
        \end{minipage}
\end{figure*}
 
As is shown in \cref{fig:GPA}, $\mathcal{P}_{pivot}$ are evenly distributed among $\mathcal{P}$.
Then the feature $\mathbf{X}_{pivot} \in \mathbb{R}^{M \times C_g}$ according to $\mathcal{P}_{pivot}$ is projected into $\mathbf{K}$ and $\mathbf{V}$ while $\mathbf{Q}$ is mapped from $\mathbf{X}_g$ according to $\mathcal{P}$:
\begin{equation}
\label{QKV}
\begin{aligned}
    \mathbf{Q} &= \varphi(\mathbf{X}_g),\\
    \mathbf{K} &= \psi(\mathbf{X}_{pivot}),\\
    \mathbf{V} &= \alpha(\mathbf{X}_{pivot}),
\end{aligned}
\end{equation}
where $\varphi$, $\psi$, $\alpha$ are MLPs, $\mathbf{Q} \in \mathbb{R}^{N \times C_g}$ and $\mathbf{K},  \mathbf{V} \in \mathbb{R}^{M \times C_g}$.
Similar to standard self-attention in \cref{standard}, the attention map $\mathbf{A} \in \mathbb{R}^{N \times M}$ is computed by the matrix multiplication of $\mathbf{Q}$ and $\mathbf{K}^\top$. Then the output $\mathbf{Y}_g \in \mathbb{R}^{N \times C_g}$ is  multiplied by   $\mathbf{A}$ and $\mathbf{V}$:
\begin{equation}
\label{Yg}
\begin{aligned}
    \mathbf{A} &= \mathbf{Q} \mathbf{K}^\top, \\
    \mathbf{Y}_g &= \mathbf{A} \mathbf{V}.
\end{aligned}
\end{equation}
Compared with standard self-attention in \cref{standard}, GPA reduces the computation complexity from $\mathcal{O}(N^2C)$ to $\mathcal{O}(MNC)$, where $M$ is determined by $SR$. Meanwhile, through the evenly distributed pivot points, the points can interact with distant points, which balances the effective receptive field and computation cost effectively. 

\subsection{Asymmetric Parallel Sturcture}
\label{ramp}
Moreover, based on the observation\cite{MaithraRaghu2021DoVT,si2022inception} that the bottom layers of the network usually tend to capture the local features such as textures, while the top layers pay more attention to the global information such as the structures of the whole scene, we design the asymmetric parallel structure which adopts the ramp channel-ratio and sampling-ratio across the stages. 

Specifically, rather than equally splitting the channel between the local and global branches,  we gradually increase the channel-ratio of the global branch from the bottom to the top.
In detail, as is shown in \cref{fig:network}, the overall network has 5 stages.
The channel-ratio of the global branch $CR_g = \frac{C_g}{C}$ is increased stage by stage, while the channel-ratio of the local branch $CR_l = \frac{C_l}{C}$ is decreased accordingly since $CR_g + CR_l =1$. This design can balance local and global information effectively across all stages.
Ablation study in \cref{table:rampcr} shows its effectiveness.

Similarly, we adopt the ramp sampling-ratio structure. Specifically, we gradually increase the sampling-ratio from lower layers to higher layers. This structure accords with the observation\cite{MaithraRaghu2021DoVT,si2022inception} and the effectiveness is verified in \cref{table:rampsr}. 
\subsection{Network Architecture}
Based on APPT Block (\cref{APPT}), we build Asymmetric Parallel Point Transformer for 3D point cloud understanding, as depicted in \cref{fig:network}. 
Following previous work\cite{zhao2020point}, we adopt the U-Net architecture for semantic segmentation. 
There are 5 stages in encoder and decoder with the block depth [2, 3, 4, 6, 3] and [1, 1, 1, 1, 1] respectively. 
The channel dimension is set as [32, 64, 128, 256, 512]. 

In every stage, we use APPT block with the local and global branches to model short-range and long-range dependencies. For the local branch, we adopt LGA with 16 neighbors. 
On the other hand, we use GPA (\cref{GPA}) with the ramp sampling-ratio structure (\cref{ramp}) as the global branch whose sampling-ratio depth is
[0, $\frac{1}{64}$, $\frac{1}{16}$, $\frac{1}{4}$, 1].
Meanwhile, we adopt the ramp channel-ratio structure (\cref{ramp}) and the channel-ratio of it is [0, $\frac{1}{8}$, $\frac{1}{8}$, $\frac{1}{4}$, 1].

Note that we utilize a simple MLP to embed the input feature in the first stage of the encoder. Similarly, in the first stage of the decoder, we use a transformation layer implemented by MLPs to process features from the encoder.

For shape classification, we utilize global average pooling to get a global feature vector and feed it into the MLPs to get the classification logits. 
Further details will be described in the supplementary material.

\section{Experiments}
\label{sec:experiments}
In this section, we evaluate the effectiveness of our proposed Asymmetric Parallel Point Transformer for different tasks. 
For 3D semantic segmentation, we use the challenging Stanford Large-Scale 3D Indoor Spaces (S3DIS) dataset\cite{armeni_cvpr16}. Besides, we adopt the commonly used ModelNet40 dataset\cite{wu2015modelnet} for 3D shape classification and ShapeNet dataset\cite{chang2015shapenet} for 3D part segmentation.

\noindent
{\bf Implementation Detail.}
For 3D semantic segmentation on S3DIS, we use SGD optimizer with momentum 0.9 and weight decay 0.0001 respectively.
Note that we use the same data augmentation strategies as Point Transformer\cite{zhao2020point} for a fair comparison. 
For 3D shape classification on ModelNet40, we use the Adam optimizer with weight decay of 0.0001.
And for 3D part segmentation on ShapeNet, we use SGD optimizer with momentum 0.9 and weight decay 0.0001 respectively.
Further details will be described in the supplementary material.

\subsection{Semantic Segmentation}
\noindent
{\bf Data and Metric.}
For 3D semantic segmentation, we conduct experiments on S3DIS\cite{armeni_cvpr16} dataset to evaluate our proposed Asymmetric Parallel Point Transformer. The S3DIS dataset contains 271 rooms in six areas from three different buildings. The points are annotated into 13 categories \eg, bookcase, floor, and board. 
Following a common protocol\cite{qi2017pointnet++,zhao2020point}, we evaluate the presented model in two approaches: (1) Area 5 is withheld during training and used for testing. (2) 6-fold cross-validation. And we adopt mean class-wise intersection over union (mIoU), mean of class-wise accuracy (mAcc), and overall point-wise accuracy (OA) as the evaluation metric for performance. 
On the other hand, we use the number of parameters and average FLOPs on Area 5 as the evaluation metric for efficiency.

\noindent
{\bf Result.}
The results are reported in \cref{tab:s3disresult} and \cref{tab:s3disresult2}. On Area 5, our proposed Asymmetric Parallel Point Transformer outperforms prior Point Transformer by 2.6/1.9 absolute percentage points in mAcc and mIoU respectively.
Moreover, our model outperforms the state-of-the-art model Stratified Transformer by 1.0/0.3 absolute percentage points in mAcc and mIoU, which verifies the effectiveness of our model. 

On the other hand, to enlarge the effective receptive field, Stratified Transformer adopts window-based attention with a stratified key sampling strategy and special memory-efficient implementation, which brings in heavy computation costs. 
Contrarily, equipped with Global Pivot Attention and parallel design, our model enlarges the effective receptive field without extra computation costs.
As shown in \cref{tab:s3disresult}, the FLOPs of our model are fewer than half of Stratified Transformer and slightly fewer than Point Transformer,
which verifies the efficiency of our model. 

For the 6-fold cross-validation, the result is presented in \cref{tab:s3disresult2}. Our model achieves 91.3\%/86.6\%/77.5\% in OA, mAcc, and mIoU which outperforms all the prior models significantly. Specifically, ours outperforms the prior state-of-the-art by 0.5, 3.6, and 2.6 absolute percentage points in mAcc and mIoU, which validates the effectiveness of long-range dependencies and
demonstrates the superiority of our proposed APPT.

\begin{table}
    \centering
    {
        \begin{footnotesize}
        \begin{tabular}{ l |   c c | c  c }
            \toprule
            	Method    & \makecell[c]{mAcc\\(\%)} & \makecell[c]{mIoU\\(\%)} & \makecell[c]{Param\\(M)} & \makecell[c]{FLOPs\\(G)} \\

            \specialrule{0em}{2pt}{0pt}
\hline
\specialrule{0em}{2pt}{0pt}
PointNet~\cite{qi2017pointnet}  & 49.0 & 41.1  & -- & --\\

PointCNN~\cite{li2018pointcnn}   & 63.9 & 57.3 & -- & --\\

PointWeb~\cite{zhao2019pointweb}   & 66.6 & 60.3 & -- & --\\

HPEIN~\cite{jiang2019hierarchical} & 68.3 & 61.9  & -- & --\\

ParamConv~\cite{wang2018deep}   & 67.0 & 58.3 & -- & --\\

SPGraph~\cite{landrieu2018large}  & 66.5 & 58.0 & -- & --\\

SegGCN~\cite{lei2020seggcn}   & 70.4 & 63.6 & -- & --\\

MinkowskiNet~\cite{choy20194d}   & 71.7 & 65.4 & -- & --\\

RepSurf-U\cite{ran2022surface}          & 68.9  &76.0 &-- &-- \\


KPConv~\cite{thomas2019kpconv}  & 72.8 & 67.1 & 150 & 334\\

PointMixer~\cite{choe2022pointmixer}  & 77.9 & 71.4 & \textbf{6.5} &-- \\

PointNext-XL~\cite{qianpointnext}  & 77.4 & 71.1& 41.6 & \textbf{84.8}  \\

FPT~\cite{ChunghyunPark2022FastPT}   & 77.3 & 70.1 & 37.9 & --  \\

PT~\cite{zhao2020point} & 76.5 & 70.4 & \underline{7.8} & 138\\

PTv2~\cite{wupoint}  & 77.9 & 71.6  & 12.8 & -- \\

ST~\cite{XinLai2022StratifiedTF}  & \underline{78.1} & \underline{72.0} & 8.0 & 280\\
\specialrule{0em}{2pt}{0pt}
\hline
\specialrule{0em}{2pt}{0pt}

APPT (Ours)   & \textbf{79.1} & \textbf{72.3} & 8.4 & \underline{126}\\
            
            \bottomrule                                   
        \end{tabular}
        \end{footnotesize}
    }    
    \vspace{0.1cm}
    \caption{Results on S3DIS Area 5 for semantic segmentation.}
    \label{tab:s3disresult}   
\vspace{-0.2cm}
\end{table}

\begin{table}
    \centering
    {
        \begin{footnotesize}
        \begin{tabular}{ l | c |  c | c }
            \toprule
            	Method   & \makecell[c]{OA}& \makecell[c]{mAcc} & \makecell[c]{mIoU}  \\

            \specialrule{0em}{0pt}{1pt}
            \hline
            \specialrule{0em}{0pt}{1pt}
            
        			PointNet~\cite{qi2017pointnet} & 78.5 & 66.2 & 47.6 \\
			RSNet~\cite{huang2018recurrent} & -- & 66.5 & 56.5 \\
			SPGraph~\cite{landrieu2018spg} & 85.5 & 73.0 & 62.1 \\
			PAT~\cite{yang2019modeling} & -- & 76.5 & 64.3 \\
			PointCNN~\cite{li2018pointcnn} & 88.1 & 75.6 & 65.4 \\
			PointWeb~\cite{zhao2019pointweb} & 87.3 & 76.2 & 66.7 \\
			ShellNet~\cite{zhang2019shellnet} & 87.1 & -- & 66.8 \\
			RandLA-Net~\cite{hu2020randla} & 88.0 & 82.0 & 70.0 \\
			KPConv~\cite{thomas2019kpconv} & -- & 79.1 & 70.6 \\
			PT~\cite{zhao2020point} & 90.2 & 81.9 & 73.5 \\
                RepSurf-U~\cite{ran2022surface} & \underline{90.8} 	&82.6	&74.3\\
                PointNext-XL~\cite{qianpointnext} & 90.3 & \underline{83.0}  &\underline{74.9} \\
            
            \specialrule{0em}{0pt}{1pt}
            \hline
            \specialrule{0em}{0pt}{1pt}
            
            APPT (Ours) & \textbf{91.3} & \textbf{86.6} & \textbf{77.5}\\
            
            \bottomrule                                   
        \end{tabular}
        \end{footnotesize}
    }    
    \vspace{0.1cm}
    \caption{Results on S3DIS 6-fold for semantic segmentation.}
    \label{tab:s3disresult2}   
\vspace{-0.4cm}
\end{table}

\subsection{Shape Classification}
\noindent
{\bf Data and Metric.}
For 3D shape classification, we experiment on the ModelNet40 dataset\cite{wu2015modelnet}. It consists of 12,311 CAD models with 40 object categories, which are divided into 9,843 models for training and 2,468 models for testing. Following previous works\cite{qi2017pointnet++,zhao2020point}, we uniformly sample 1,024 points from the mesh faces for each model and rescale the points to fit the unit sphere.
For the evaluation metric, we use class-average accuracy (mAcc) and overall accuracy (OA).

\noindent
{\bf Result.}
The results are reported in \cref{tab:modelnet40result}. 
Our model achieves 91.9\% in mAcc which surpasses all the previous ones significantly.
Notably, it outperforms Point Transformer by 1.3 and 0.2 absolute percentage points in mAcc and OA.
\begin{table}
    \centering
    {
        \begin{footnotesize}
        \begin{tabular}{ l |  c | c }
            \toprule
            	Method  & \makecell[c]{mAcc} & \makecell[c]{OA}  \\

            \specialrule{0em}{0pt}{1pt}
            \hline
            \specialrule{0em}{0pt}{1pt}
            
        		3DShapeNets~\cite{wu2015modelnet}  & 77.3 & 84.7 \\
		PointNet~\cite{qi2017pointnet}  & 86.2 & 89.2 \\
		PointNet++~\cite{qi2017pointnet++}  & -- & 91.9 \\
		PointConv~\cite{wu2019pointconv}  & -- & 92.5 \\
		Point2Sequence~\cite{liu2019point2sequence}  & 90.4 & 92.6 \\
		KPConv~\cite{thomas2019kpconv} & -- & 92.9 \\
		DGCNN~\cite{wang2019dynamic}        & 90.2          & 92.9          \\
    PointASNL~\cite{yan2020pointasnl}      &--             &92.9 \\
    PCT~\cite{guo2021pct}               & --             & 93.2          \\
    CPT~\cite{kaul2022convolutional} & 90.6 & 93.9 \\
    Paconv~\cite{xu2021paconv} & --             & 93.6          \\
    PointNext~\cite{qianpointnext}  & 91.1            & 94.0                \\
    PointMixer~\cite{choe2022pointmixer}  & 91.4             & 93.6          \\
    RepSurf-U~\cite{ran2022surface}  & 91.4             & \textbf{94.4}          \\
		PT~\cite{zhao2020point}  & 90.6 & 93.7 \\
  PTv2~\cite{wupoint}  & 91.6 & 94.2 \\
            
            \specialrule{0em}{0pt}{1pt}
            \hline
            \specialrule{0em}{0pt}{1pt}
            
            APPT (Ours)  & \textbf{91.9} & 93.9 \\
            
            \bottomrule                                   
        \end{tabular}
        \end{footnotesize}
    }    
    \vspace{0.1cm}
    \caption{Results on ModelNet40 for shape classification.}
    \label{tab:modelnet40result}   
\vspace{-0.1cm}
\end{table}

\subsection{Part Segmentation}
\noindent
{\bf Data and Metric.}
For 3D part segmentation, we evaluate on the ShapeNet dataset\cite{chang2015shapenet}, which contains 16,880 models from 16 shape categories. 14,006 models are used for training and 2874 models are used for testing. The annotations are divided into 2 to 6 parts and the whole dataset has 50 different parts.
For evaluation metrics, we report category mIoU and instance mIoU. 

\noindent
{\bf Result.}
The results are reported in \cref{tab:exp_shapenet}. Our model achieves 85.2\%/87.1\% in category mIoU and instance mIoU, which surpasses all the prior works. 
Specifically, it outperforms the prior state-of-the-art Stratified Transformer by 0.1 and 0.5 absolute percentage points in category mIoU and instance mIoU.

\begin{table}
    \centering
    {
        \begin{footnotesize}
        \begin{tabular}{ l | c |  c }
            \toprule
            Method & \makecell[c]{Cat. mIoU} & \makecell[c]{Ins. mIoU} \\

            \specialrule{0em}{0pt}{1pt}
            \hline
            \specialrule{0em}{0pt}{1pt}
            
            PointNet~\cite{qi2017pointnet} & 80.4 & 83.7 \\
            PointNet++~\cite{qi2017pointnet++} & 81.9 & 85.1 \\
            PCNN~\cite{atzmon2018point} & 81.8 & 85.1 \\
            SpiderCNN~\cite{xu2018spidercnn} & 82.4 & 85.3 \\
            SPLATNet~\cite{su2018splatnet} & 83.7 & 85.4 \\ 
            DGCNN~\cite{wang2019dynamic} & 82.3 & 85.2 \\
            SubSparseCNN~\cite{3DSemanticSegmentationWithSubmanifoldSparseConvNet} & 83.3 & 86.0 \\
            PointCNN~\cite{li2018pointcnn} & 84.6 & 86.1 \\
            Point2Sequence~\cite{liu2019point2sequence} & - & 85.2 \\ 
            PVCNN~\cite{liu2019pvcnn} & - & 86.2 \\
            RS-CNN~\cite{liu2019relation} & 84.0 & 86.2 \\
            KPConv~\cite{thomas2019kpconv} & 85.0 & 86.2 \\
            InterpCNN~\cite{mao2019interpolated} & 84.0 & 86.3 \\ 
            DensePoint~\cite{liu2019densepoint} & 84.2 & 86.4 \\ 
            PAConv~\cite{xu2021paconv} & 84.6 & 86.1 \\
            PT~\cite{zhao2020point} & 83.7 & 86.6\\
            ST~\cite{XinLai2022StratifiedTF} & 85.1 & 86.6 \\
            
            \specialrule{0em}{0pt}{1pt}
            \hline
            \specialrule{0em}{0pt}{1pt}
            
            APPT (Ours) & \textbf{85.2} & \textbf{87.1} \\
            
            \bottomrule                                   
        \end{tabular}
        \end{footnotesize}
    }    
    \vspace{0.1cm}
    \caption{Results on ShapeNet for part segmentation.}
    \label{tab:exp_shapenet}   
\vspace{-0.3cm}
\end{table}

\begin{figure*}
	\centering
     \begin{minipage}  {0.19\linewidth}
        \centering
        \includegraphics [width=0.75\linewidth,height=0.75\linewidth]
        {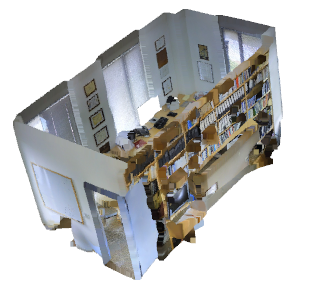}
    \end{minipage}     
    \begin{minipage}  {0.19\linewidth}
        \centering
        \includegraphics [width=0.8\linewidth,height=0.8\linewidth]
        {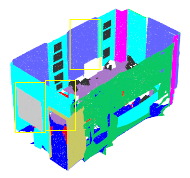}
    \end{minipage}      
     \begin{minipage}  {0.19\linewidth}
        \centering
        \includegraphics [width=0.8\linewidth,height=0.8\linewidth]
        {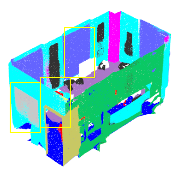}
    \end{minipage} 
     \begin{minipage}  {0.19\linewidth}
        \centering
        \includegraphics [width=0.8\linewidth,height=0.8\linewidth]
        {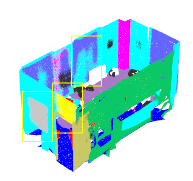}
    \end{minipage} 
    \begin{minipage}  {0.19\linewidth}
        \centering
        \includegraphics [width=0.8\linewidth,height=0.8\linewidth]
        {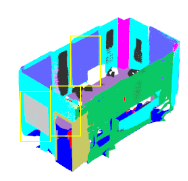}
    \end{minipage}      
	 
    \begin{minipage}  {0.19\linewidth}
        \centering
        \includegraphics [width=0.75\linewidth,height=0.75\linewidth]
        {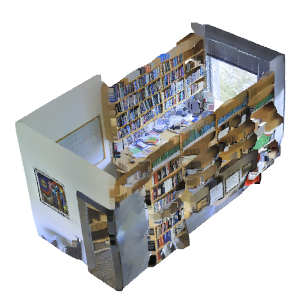}
    \end{minipage}   
    \begin{minipage}  {0.19\linewidth}
        \centering
        \includegraphics [width=0.8\linewidth,height=0.8\linewidth]
        {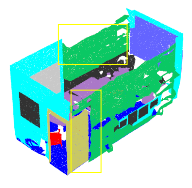}
    \end{minipage}
     \begin{minipage}  {0.19\linewidth}
        \centering
        \includegraphics [width=0.8\linewidth,height=0.8\linewidth]
        {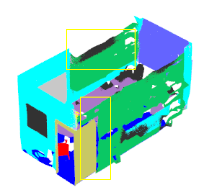}
    \end{minipage} 
     \begin{minipage}  {0.19\linewidth}
        \centering
        \includegraphics [width=0.8\linewidth,height=0.8\linewidth]
        {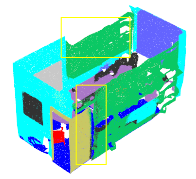}
    \end{minipage} 
    \begin{minipage}  {0.19\linewidth}
        \centering
        \includegraphics [width=0.8\linewidth,height=0.8\linewidth]
        {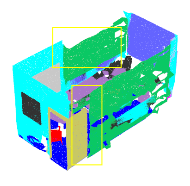}
    \end{minipage}      
    
	 
     \begin{minipage}  {0.19\linewidth}
        \centering
        \includegraphics [width=0.75\linewidth,height=0.75\linewidth]
        {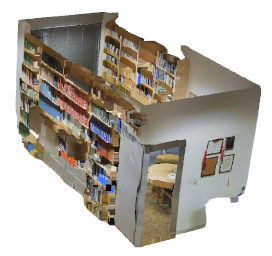}\\\footnotesize{Input}
    \end{minipage} 
    \begin{minipage}  {0.19\linewidth}
        \centering
        \includegraphics [width=0.8\linewidth,height=0.8\linewidth]
        {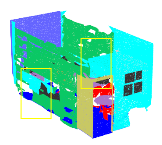}\\\footnotesize{Ground Truth}
    \end{minipage} 
     \begin{minipage}  {0.19\linewidth}
        \centering
        \includegraphics [width=0.8\linewidth,height=0.8\linewidth]
        {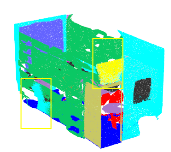}\\\footnotesize{PT}
    \end{minipage} 
     \begin{minipage}  {0.19\linewidth}
        \centering
        \includegraphics [width=0.8\linewidth,height=0.8\linewidth]
        {fig/vis/o6st.png}\\\footnotesize{ST}
    \end{minipage} 
    \begin{minipage}  {0.19\linewidth}
        \centering
        \includegraphics [width=0.8\linewidth,height=0.8\linewidth]
        {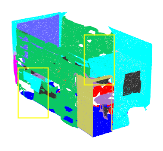}\\\footnotesize{APPT}
    \end{minipage}      
	 
	 \vspace{0.1cm}
	 
    \begin{minipage}  {0.04\linewidth}
        \centering
        \includegraphics [width=0.5\linewidth,height=0.5\linewidth]
        {fig/color/board.png}
    \end{minipage}\footnotesize board
    \begin{minipage}  {0.04\linewidth}
        \centering
        \includegraphics [width=0.5\linewidth,height=0.5\linewidth]
        {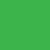}
    \end{minipage}\footnotesize bookcase
    \begin{minipage}  {0.04\linewidth}
        \centering
        \includegraphics [width=0.5\linewidth,height=0.5\linewidth]
        {fig/color/chair.png}
    \end{minipage}\footnotesize chair
    \begin{minipage}  {0.04\linewidth}
        \centering
        \includegraphics [width=0.5\linewidth,height=0.5\linewidth]
        {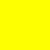}
    \end{minipage}\footnotesize ceiling
    \begin{minipage}  {0.04\linewidth}
        \centering
        \includegraphics [width=0.5\linewidth,height=0.5\linewidth]
        {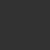}
    \end{minipage}\footnotesize clutter
    \begin{minipage}  {0.04\linewidth}
        \centering
        \includegraphics [width=0.5\linewidth,height=0.5\linewidth]
        {fig/color/bookcase.png}
    \end{minipage}\footnotesize door
    \begin{minipage}  {0.04\linewidth}
        \centering
        \includegraphics [width=0.5\linewidth,height=0.5\linewidth]
        {fig/color/floor.png}
    \end{minipage}\footnotesize floor
    \begin{minipage}  {0.04\linewidth}
        \centering
        \includegraphics [width=0.5\linewidth,height=0.5\linewidth]
        {fig/color/table.png}
    \end{minipage}\footnotesize table
    \begin{minipage}  {0.04\linewidth}
        \centering
        \includegraphics [width=0.5\linewidth,height=0.5\linewidth]
        {fig/color/wall.png}
    \end{minipage}\footnotesize wall
    \begin{minipage}  {0.04\linewidth}
        \centering
        \includegraphics [width=0.5\linewidth,height=0.5\linewidth]
        {fig/color/window.png}
    \end{minipage}\footnotesize window
	 
    \vspace{0.2cm}     
    \caption{Visual comparison between Point Transformer (PT), Stratified Transofrmer (ST) and our proposed Asymmetric Parallel Point Transformer (APPT) on S3DIS dataset.}
    \label{fig:vis_comparison}
    \vspace{-0.3cm}
\end{figure*}

\subsection{Ablation Study}
To verify the effectiveness of each component in our Asymmetric Parallel Point Transformer, we conduct a number of ablation studies. The results are reported on the semantic segmentation S3DIS dataset, tested on Area 5.

\noindent
{\bf Local and global module.}
To evaluate the importance of each module, we conduct ablation studies in \cref{tab:ppt} with local module only and global module only.
Observably from Exp.\uppercase\expandafter{\romannumeral1} to Exp.\uppercase\expandafter{\romannumeral4}, 
equipped with the global module, the model achieves better mIoU than the local module only both parallelly and serially, which indicates the effectiveness of the global module to enlarge the receptive field.
On the other hand, Exp.\uppercase\expandafter{\romannumeral2} also shows the great importance of local module
to model short-range dependencies.

\noindent
{\bf Parallel design.}
The parallel design is composed of local and global branches to model short-range and long-range dependencies respectively. 
To evaluate the effectiveness of the design, we compare it with 
the serial design which puts the global module after the local module in every block.   
Observably from Exp.\uppercase\expandafter{\romannumeral3} and Exp.\uppercase\expandafter{\romannumeral4}, 
our proposed parallel design with 72.3\% mIoU significantly outperforms the serial approach with 70.9\% mIoU, which demonstrates its capability to effectively incorporate local and global features.

\noindent
{\bf Global branch.}
To verify our proposed GPA in the global branch,  we compare it with other global modules, \ie Stratified Transformer block (ST)\cite{XinLai2022StratifiedTF}, Squeeze-Extraction (SE) \cite{JieHu2018SqueezeandExcitationN} and Global Context (GC) \cite{YueCao2019GCNetNN}.

As shown in \cref{tab:ppt} (from Exp.\uppercase\expandafter{\romannumeral3} to Exp.\uppercase\expandafter{\romannumeral10}), no matter parallel and serial, our GPA is more expressive than other global modules. Specifically for parallel approach (Exp.\uppercase\expandafter{\romannumeral4}, \uppercase\expandafter{\romannumeral8}, \uppercase\expandafter{\romannumeral9} and \uppercase\expandafter{\romannumeral10}), the performance gap between our proposed GPA and other modules are significant. 
Particularly, our GPA outperforms ST by a large margin, which verifies its great power to model long-range dependencies and enlarge the effective receptive field.
Although SE and GC have a good ability to model long-range dependencies in 2D image understanding, our method is more suitable for 3D point cloud understanding and achieves better performance. 

\begin{table}[!t]
    \centering
    {
        \begin{footnotesize}
        \begin{tabular}{ c |c  c| c | c c c c | c  }
            \toprule
            ID & Parallel & Serial & LGA  & GPA & ST & SE & GC  & mIoU  \\

            \specialrule{0em}{0pt}{1pt}
            \hline
            \specialrule{0em}{0pt}{1pt}
            
            \uppercase\expandafter{\romannumeral1} & & & \checkmark& & & & & 70.4  \\ 

            \uppercase\expandafter{\romannumeral2} & & && \checkmark& & & &  67.0  \\ 

            \uppercase\expandafter{\romannumeral3} & & \checkmark& \checkmark &  \checkmark & & & & 70.9  \\ 
            
            \uppercase\expandafter{\romannumeral4} & \checkmark& & \checkmark & \checkmark& & & &  \textbf{72.3}\\

            \specialrule{0em}{0pt}{1pt}
            \hline
            \specialrule{0em}{1pt}{0pt}

            \uppercase\expandafter{\romannumeral5} & &\checkmark & \checkmark &   & \checkmark& && 70.3  \\ 
            
            \uppercase\expandafter{\romannumeral6} & &\checkmark & \checkmark & &  & \checkmark& & 70.5  \\ 
            
            \uppercase\expandafter{\romannumeral7} & &\checkmark & \checkmark & &  &  & \checkmark& 70.4  \\
            
            \uppercase\expandafter{\romannumeral8} & \checkmark& & \checkmark &  & \checkmark& & &  71.0  \\
            
            \uppercase\expandafter{\romannumeral9} & \checkmark& &\checkmark&  &  & \checkmark & &  70.7  \\
            
            \uppercase\expandafter{\romannumeral10} & \checkmark& &\checkmark&  & & & \checkmark & 70.8 \\ 
            
            
            \bottomrule                                   
        \end{tabular}
        \end{footnotesize}
    }    
    \vspace{0.1cm}
    \caption{Ablation study for parallel design and global branch.
    \textbf{Parallel}: Design of parallel branches.
    \textbf{Serial}: Design of serial blocks. 
    \textbf{LGA}: Local Group Attention in local branch. 
    \textbf{GPA}: Global Pivot Attention.
    \textbf{ST}: Stratified Transformer block\cite{XinLai2022StratifiedTF}.
    \textbf{SE}: Squeeze-Extraction\cite{JieHu2018SqueezeandExcitationN}. 
    \textbf{GC}: Global Context\cite{YueCao2019GCNetNN}. }
    \label{tab:ppt}   
\vspace{-0.2cm}
\end{table}

\noindent
{\bf Ramp channel-ratio.}
Based on the investigations\cite{MaithraRaghu2021DoVT,si2022inception} that the lower layer requires more local information and the higher layer pays more attention to global dependencies for the Transformer network, we design the ramp channel-ratio structure that increases the channel-ratio of the global from the bottom to the top stage. To verify its effectiveness, we investigate different strategies for channel splitting in \cref{table:rampcr}. We can clearly see that the strategy of $C_g/C \uparrow$ outperforms others.
The improvement accords with the observation and demonstrates the effectiveness of our design.

\begin{table}[!t]
    \centering
    {
        \begin{footnotesize}
        \begin{tabular}{  c | c | c}
            \toprule
            Channel-ratio  & mIoU & GFLOPs  \\
            
            \specialrule{0em}{0pt}{1pt}
            \hline
            \specialrule{0em}{0pt}{1pt}
            
            $C_g/C\uparrow, C_l/C\downarrow$ & \textbf{72.3} & 126  \\

            $C_g/C = C_l/C$ & 71.1 & 127 \\

            $C_g/C\downarrow, C_l/C\uparrow$ & 71.0 & \textbf{122} \\
            
            \bottomrule                                   
        \end{tabular}
        \end{footnotesize}
    }    
    \vspace{0.1cm}
    \caption{Ablation for ramp channel-ratio. $C_g$ is the channel of the global branch and $C$ is the total channel. 
    }
    \label{table:rampcr}
\end{table}

\noindent
{\bf Ramp sampling-ratio.}
Similar to the ablation study of ramp channel-ratio structure, we perform experiments on different strategies for sampling-ratio in \cref{table:rampsr}. The result shows that the strategy of $M/N \uparrow$ outperforms others,
which verifies the effectiveness of the structure.

\begin{table}[!t]
    \centering
    {
        \begin{footnotesize}
        \begin{tabular}{  c | c | c  }
            \toprule
            Sampling-ratio  & mIoU & GFLOPs  \\
            
            \specialrule{0em}{0pt}{1pt}
            \hline
            \specialrule{0em}{0pt}{1pt}
                       
            $M/N\uparrow$ & \textbf{72.3} & 126  \\

            $M/N$ fixed & 71.6 & \textbf{125} \\

            $M/N\downarrow$ & 71.2 & 128 \\

            
            \bottomrule                                   
        \end{tabular}
        \end{footnotesize}
    }    
    \vspace{0.1cm}
    \caption{Ablation for ramp sampling-ratio. $M$ is the number of pivot points and $N$ is the number of all the points. 
    }
    \label{table:rampsr}
\vspace{-0.2cm}
\end{table}

\noindent
{\bf Fusion strategy.}
We conduct an ablation study on fusion strategy for the output of local and global branches and examine three strategies: simple concatenation, concatenation with Squeeze-Extraction\cite{JieHu2018SqueezeandExcitationN}, and summation with MLPs.
Observably from \cref{table:fusion}, the simple concatenation outperforms other fusion strategies, demonstrating the ability to integrate local and global components more effectively.  
\begin{table}[!t]
    \vspace{0.1cm}
    \centering
    {
        \begin{footnotesize}
        \begin{tabular}{  c | c | c |c  }
            \toprule
            Fusion & Concat & SE & Sum  \\
            
            \specialrule{0em}{0pt}{1pt}
            \hline
            \specialrule{0em}{0pt}{1pt}
            
            mIoU & \textbf{72.3} &  70.7 & 71.5  \\

            
            \bottomrule                                   
        \end{tabular}
        \end{footnotesize}
    }    
    \vspace{0.1cm}
    \caption{Ablation for fusion strategy. \textbf{Concat}: Simple concatenation. \textbf{SE}: Concatenation with Squeeze-Extraction\cite{JieHu2018SqueezeandExcitationN}. \textbf{Sum}:  Summation with MLPs.}
    \label{table:fusion}
\vspace{-0.3cm}
\end{table}

\noindent
{\bf Visual comparison.}
As shown in \cref{fig:vis_comparison}, we visually compare Point Transformer\cite{zhao2020point}, Stratified Transformer\cite{XinLai2022StratifiedTF} and our proposed Asymmetric Parallel Point Transformer on S3DIS dataset. We can clearly see that our method outperforms priors with the ability to recognize the objects highlighted with the yellow boxes, benefiting from its great power to model long-range dependencies. 
\section{Conclusion}
\label{sec:conclusion}
In this paper, we propose Asymmetric Parallel Point Transformer (APPT), a powerful framework for 3D point cloud understanding. 
To enlarge the effective receptive field, we introduce Global Pivot Attention as the global branch.
Moreover, we design the Asymmetric Parallel structure to effectively integrate local and global information.
Benefiting from the ability to capture long-range dependencies, our model outperforms prior transformer-based networks and achieves state-of-the-art performance on semantic segmentation, shape classification, and part segmentation benchmarks. 

{\small
\bibliographystyle{ieee_fullname}
\bibliography{egbib}
}

\end{document}